\documentclass{article}
\usepackage{smlia2024}
\usepackage{multicol}
\usepackage[utf8]{inputenc} 
\usepackage[T1]{fontenc}    
\usepackage{hyperref}       
\usepackage{url}            
\usepackage{booktabs}       
\usepackage{amsmath} 
\usepackage{comment} 
\usepackage{listings} 
\usepackage{ascmac} 
\usepackage{amsfonts}       

\usepackage{nicefrac}       
\usepackage{microtype}      
\usepackage{cleveref}       
\usepackage{lipsum}         
\usepackage{graphicx}
\usepackage{doi}
\usepackage{wrapfig}
\usepackage{float}

\title{Estimate epidemiological parameters given partial observations based on algebraically observable PINNs}

\renewcommand{\shorttitle}{Estimate Epidemiological Parameters given Partial Observations based on Algebraically Observable PINNs}

\date{}

\newif\ifuniqueAffiliation
\uniqueAffiliationtrue

\ifuniqueAffiliation 
\author{%
	Mizuka Komatsu \\
	  Graduate School of System Informatics\\
	Kobe University\\
	1-1 Rokkodai-cho, Nada-ku, Kobe 657-8501, Japan \\
	\texttt{m-komatsu@bear.kobe-u.ac.jp} 
}
\else

\usepackage{authblk}

\fi

\hypersetup{
pdftitle={A template for the arxiv style},
pdfsubject={q-bio.NC, q-bio.QM},
pdfauthor={David S.~Hippocampus, Elias D.~Striatum},
pdfkeywords={First keyword, Second keyword, More},
}

\begin{document}
\maketitle
\begin{abstract}
In this study, we considered the problem of estimating epidemiological parameters based on physics-informed neural networks (PINNs). In practice, not all trajectory data corresponding to the population estimated by epidemic models can be obtained, and some observed trajectories are noisy. Learning PINNs to estimate unknown epidemiological parameters using such partial observations is challenging. Accordingly, we introduce the concept of algebraic observability into PINNs. The validity of the proposed PINN, named as an algebraically observable PINNs, in terms of estimation parameters and prediction of unobserved variables, is demonstrated through numerical experiments.
\end{abstract}

\begin{multicols}{2}

\section{Introduction}\label{sec:intro}
In this study, we investigated the problem of estimating epidemiological parameters. Epidemiological parameters refer to the parameters that appear in epidemiological models such as the SIR and SEIR models\cite{inaba, kuniya} where populations are assumed to be divided into subgroups depending on infectious states and the transition among the subgroups. The values of these parameters are significant in investigating trends in infectious diseases such as COVID-19\cite{kuniya}. 
In this study, we focused on the problem of estimating epidemiological parameters based on Physics-Informed Neural Networks, (PINNs)\cite{pinn}.

PINNs are Deep Neural Networks used to predict phenomena by incorporating prior knowledge to adhere to certain governing equations.
In this study, we assume that the governing equations are ordinary differential equations (ODEs) in which the dependent and independent variables are represented as $x(t; \theta) \in \mathbb{R}^N$ and $t$ respectively. $\theta \in \mathbb{R}^M$ denotes the parameters of the ODEs, such as the epidemiological parameters. We denote the data observed at $t$ as $x_{\mathrm{d}}(t)$. In this case, the input and output of the PINNs correspond to $t$ and $x(t)$, respectively.  PINNs are trained such that the output of PINNs $x_{\mathrm{nn}}$ given $w, \theta$ approximates $x_{\mathrm{data}}$ satisfies the ODEs as much as possible, where $w \in \mathbb{R}^L$ denotes the weight parameters of the PINNs. For this purpose, the loss function $L(x_\mathrm{nn}, x_\mathrm{d}; w, \theta)$ is defined as the sum of the error term on the data $\lambda_{\mathrm{data}}L_{\mathrm{data}}(x_\mathrm{nn}, x_\mathrm{d}; w, \theta),$ the error term of the governing equations $\lambda_{\mathrm{eq}}L_{\mathrm{eq}}(x_\mathrm{nn}, x_\mathrm{d}; w)$ and the error term of the initial conditions$ \lambda_{\mathrm{init}}L_{\mathrm{init}}(x_\mathrm{nn}, x_\mathrm{d}; w)$
where $\lambda_{\mathrm{data}}, \lambda_{\mathrm{eq}}, \lambda_{\mathrm{init}}$ are constants representing weights of each term. 
If $\theta$ is known, the learning of the PINNs is reduced to minimize $L(x_\mathrm{nn}, x_\mathrm{d}; w, \theta)$ with respect to $w$, which is known as forward problem. 
If $\theta$ includes unknown parameters, it is reduced to minimizing $L(x_\mathrm{nn}, x_\mathrm{d}; w, \theta)$ with respect to $w$ and $\theta$, which is known as the inverse problem. 
In practice, not all the trajectories corresponding to each subject can be obtained, or some of the observed trajectories are noisy because of the limited testing capacity for infection \cite{mpinn}. However, learning vanilla PINNs to estimate unknown epidemiological parameters using partial observations is challenging \cite{mpinn}. 
Motivated by this, in this paper, we propose a method for parameter estimation and population trajectory prediction using PINNs.

\vspace{-16pt}
\paragraph{Target scenario}
In this study, we consider the SEIR model \cite{inaba} as an example of an epidemic model:
\vspace{-12pt}
\begin{align}
\begin{split}
&\dot{S} = -{\beta}{S}I,\quad \dot{E} = {\beta}{S}{I}-{\epsilon}{E},\quad \dot{I} = {\epsilon}{E}-{\gamma}{I}, \\
&\dot{R} = {\gamma}{I}, \quad S + E + I + R = 1.
\end{split}\label{eq:seir}
\end{align}
The SEIR model assumes that a population is divided into four subgroups. $S$, $E$, $I$ and $R$ denote the population ratios of each subgroup, that is, susceptible, exposed, infectious, and removed, respectively.
$S, E, I, R$ depend on time $t$ and the variables with dots represent the derivatives of the variables with respect to $t$.
The transitions are described in \eqref{eq:seir}.
The total population is assumed to be constant, as represented in the last equation of \eqref{eq:seir}. 
$\beta$, $\epsilon$, $\gamma$ denote infection, onset, and removal rates, respectively. 
Throughout this study, these epidemiological parameters are assumed to be constant, and the initial values of \eqref{eq:seir} are assumed to be known. In addition, only the trajectory of the infectious population rate is assumed to be available, and $\beta, \gamma$ are known, whereas $\epsilon$ is unknown and thus estimated.

\vspace{-10pt}
\paragraph{Loss function of vanilla PINNs}
In our target scenario, $(S, E, I, R)$ and $(\beta, \epsilon, \gamma)$ correspond to $x$ and $\theta$, respectively. 
In the same manner as in Section \ref{sec:intro}, the observed data are denoted as $I_\mathrm{d}$ and the outputs of the PINN are represented as $S_\mathrm{nn}, E_\mathrm{nn}, I_\mathrm{nn}, R_\mathrm{nn}$. 
For the vanilla PINN, $L_{\mathrm{d}}$ given a full observation is defined as the mean of 
\begin{align*}
\begin{split}
&C_{\mathrm{S}} {\left(S_\mathrm{d}(t_i)- S_\mathrm{nn}(t_i)\right)}^2
+ C_{\mathrm{E}}{\left(E_\mathrm{d}(t_i)- E_\mathrm{nn}(t_i)\right)}^2\\
&+ C_{\mathrm{I}}{\left(I_\mathrm{d}(t_i)- I_\mathrm{nn}(t_i)\right)}^2
+ C_{\mathrm{R}} {\left(R_\mathrm{d}(t_i)- R_\mathrm{nn}(t_i)\right)}^2
\end{split}
\end{align*}
for $i = 1,\ldots,n$ where $C_{\mathrm{S}}, \ldots, C_{\mathrm{R}}$ and $n$ represent constants and the number of data points, respectively.
In our target scenario, we set $(C_{\mathrm{S}}, C_{\mathrm{E}}, C_{\mathrm{I}}, C_{\mathrm{R}}) = (0, 0, 1, 0)$ for the vanilla PINN. 
See Appendix \ref{sec:loss} for the definition of $L_{\mathrm{eq}}, L_{\mathrm{init}}$ in the target scenario. 

\vspace{-8pt}
\section{The algebraic observability of SEIR model}\label{sec:AO} 
\vspace{-10pt}
To overcome the difficulty in learning PINNs given partial observations, we introduced an algebraic version of observability \cite{meshkat, kalman}. Intuitively, a state variable $x$ is algebraically observable if and only if there exists a polynomial equation in which the variables are $x$, the observed variables, and the derivatives of the observed variables in the set of polynomial equations obtained through differential and algebraic manipulations of the state space models. 
We explain this concept using our target scenario, in which $I$ is assumed to be observed and $S, E, R$ are assumed to be unobserved. $E$ is algebraically observable. 
This is because the equation $\dot{I} = \epsilon E - \gamma I$ is a polynomial equation of $E$, $I$, and $\dot{I}$ regarding derivatives as independent of the variables.
In general, to obtain polynomial equations to show the algebraic observability of a state variable $x$, variables and derivatives except for $x$, observed variables and derivatives of observed variables have to be eliminated through differential and algebraic manipulations of the state space models. 
According to \cite{meshkat}, the elimination process can be reduced to eliminate a certain set of variables from a finite set of polynomial equations regarding the derivatives of variables that are independent of the variables. This problem can be solved by using the Gr\"{o}bner basis\cite{iva}. 
Furthermore, by using computer algebra software, e.g., Singular\cite{singular}, the process can be automated. 
See \ref{sec:meshkat} for further details. 
Consequently, noting that $R = 1-(S+E+I)$, all the unobserved variables in our target scenario can be recovered from $I$ and their derivatives as follows:
\vspace{-3pt}
\begin{align}
\begin{split}
    &S = \frac{\ddot{I} + (\epsilon + \gamma)\dot{I} + \epsilon\gamma{I}}{\beta\epsilon{I}}, \quad E = \frac{\dot{I} + \gamma{I}}{\epsilon}.
    \end{split}\label{eq:aoeq}
\end{align}

\section{Algebraically Observable PINNs}\label{sec:prop}

\vspace{-13pt}
In this section, we propose modified PINNs named as the algebraically obvservable PINNs.
Considering that learning the vanilla PINN with full observation is easier than with partial observation, we propose using polynomial equations that determine the algebraic observability to generate data corresponding to unobserved variables and set $(C_{\mathrm{S}}, C_{\mathrm{E}}, C_{\mathrm{I}}, C_{\mathrm{R}})$ as $(1, 1, 1, 1)$. 
In our target scenario, we used \eqref{eq:aoeq} to estimate the data for $S, E, R$.
In the following, the baseline method refers to the learning framework of the vanilla PINN with partial observation, $(C_{\mathrm{S}}, C_{\mathrm{E}}, C_{\mathrm{I}}, C_{\mathrm{R}}) = (0, 0, 1, 0)$, as an inverse problem. 
Both for the baseline and proposed method, we set $(\lambda_{\mathrm{data}}, \lambda_{\mathrm{eq}}, \lambda_{\mathrm{init}})$ as $(1, 1, 1)$.

According to \eqref{eq:aoeq}, to estimate these data, unknown values of $\dot{I}, \ddot{I}$ at the observation points and $\epsilon$ are required. For simplicity, we assume that these values are available in advance.
In the numerical experiment, we implemented a learning framework for algebraically observable PINNs with an unknown $\epsilon$ by introducing a Gaussian process-based Bayesian-optimization (GP-BO)\cite{gpbo, hporev} as an outer loop. The GP-BO is often applied to hyperparameter optimization problems\cite{hpo}. 
In our context, $\epsilon$ required to generate data corresponding to $S, E, R$ is regarded as a hyperparameter of algebraically observable PINNs. 
Specifically, GP-BO based on Expected Improvement \cite{EI, hpo} is performed to estimate $\epsilon$ such that the test error of the the algebraically observable PINNs is as much as small. 
We regard the minimal values of GP-BO as the estimated values of $\epsilon$, denoted as $\hat{\epsilon}$. The learned algebraically observable PINN, given the unobserved data generated using $\hat{\epsilon}$ is used for predicting the trajectories of $S, E, I, R$. 
For comparison, we employed the baseline method with the GP-BO such of which result is regarded as an initial estimate of $\epsilon$ for the vanilla PINN as an inverse problem. 
See Appendix D 
for the details of the numerical experiment.
\vspace{-12pt}

\begin{figure}[H]
	\begin{center}
\includegraphics[width=0.4\textwidth]{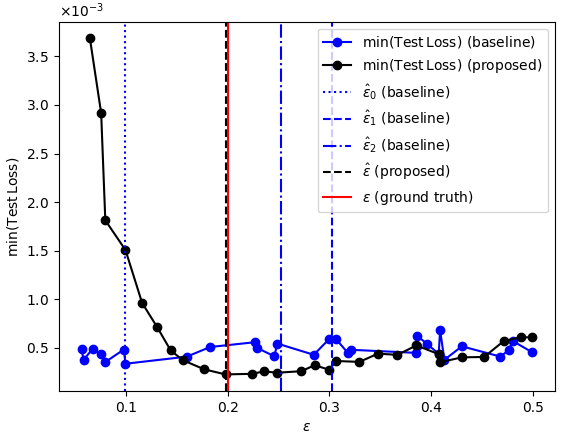}
	\end{center}
	\caption{The values of the loss function of the GP-BO.}
	\label{fig:gp}
\end{figure}

\begin{figure}[H]
	\begin{center}
\includegraphics[width=0.4\textwidth]{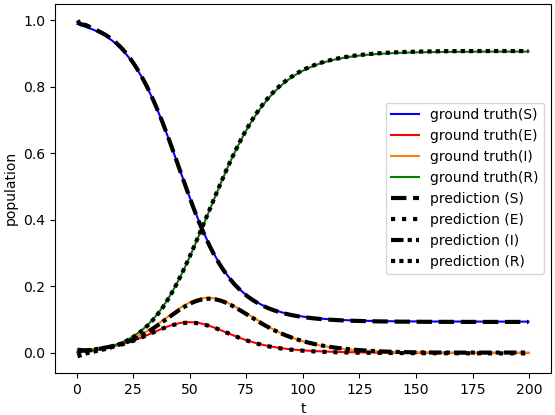}
	\end{center}
	\caption{Comparison of trajectories (Ground truth versus prediction by the proposed method.)}
	\label{fig:pred-prop}
\end{figure}

\begin{figure}[H]
	\begin{center}
\includegraphics[width=0.4\textwidth]{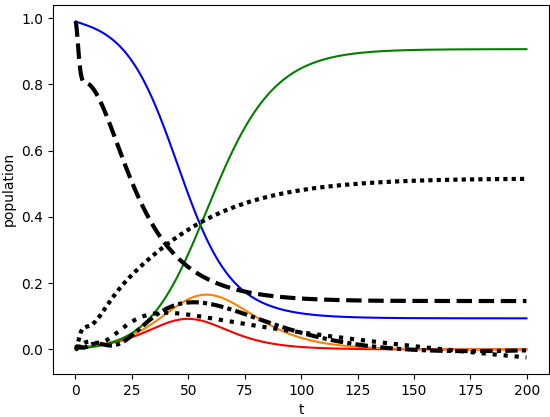}
	\end{center}
	\caption{Comparison of trajectories (Ground truth versus prediction by the baseline method at 2000 epoch) The same legend as Figure \ref{fig:pred-base-2000} is applied.}
	\label{fig:pred-base-2000}
\end{figure}

\begin{figure}[H]
	\begin{center}
\includegraphics[width=0.4\textwidth]{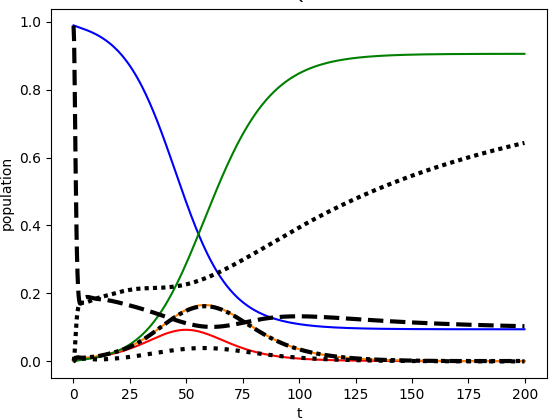}
	\end{center}
	\caption{Comparison of trajectories (Ground truth versus prediction by the baseline method at 26000 epoch.) The same legend as Figure \ref{fig:pred-base-26000} is applied.}
	\label{fig:pred-base-26000}
\end{figure}

\paragraph{Results} In each GP-BO iteration, the minimum value of the test error for learning the PINNs was evaluated. See Appendix E 
for the details.
The values of the loss function of GP-BO are shown in Figure \ref{fig:gp}. The estimated value of $\epsilon$ obtained using the proposed method $\hat{\epsilon}$ was 0.198. The absolute error is $2\times 10^{-3}$.
The predictions of $S, E, I, R$ by the algebraically observable PINN are shown in Figure \ref{fig:pred-prop}, which shows good fits not only for the observed $I$ but also for the unobserved $S, E, R$. 
For the baseline method, the initial estimate of $\epsilon$
, denoted as $\hat{\epsilon}_0$, was estimated to be 0.099. Given $\hat{\epsilon}_0$, we confirmed the estimated values of $\epsilon$ at epochs of 2000 and 26000, where the test loss was minimal ($\hat{\epsilon}_1$), and the training loss was minimal ($\hat{\epsilon}_2$). 
As shown in Figure \ref{fig:gp}, both $\hat{\epsilon}_1 = 0.303$ and $\hat{\epsilon}_2 = 0.252$ had larger absolute errors than those of the proposed method. The predictions of $S, E, I, R$ by the PINN in the baseline method are shown in Figures \ref{fig:pred-base-2000} and \ref{fig:pred-base-26000}. 

\section*{Acknowledgments}
\vspace{-10pt}
This work is supported by JST ACT-X Grant JPMJAX22A7, JSPS KAKENHI Grand Number 22K21278 and 24K16963.

\bibliographystyle{plain}

\newpage
\section*{Appendix A: Definition of $L_{\mathrm{eq}}$ in our target scenario}\label{sec:loss}
$L_{\mathrm{eq}}$ in our target scenario is defined as
\begin{align*}
\begin{split}
\frac{1}{n}\sum_{i=1}^n
&\left\{ {\left(\dot{S}_{\mathrm{nn}}(t_i)+ {\beta}{S_{\mathrm{nn}}(t_i)}I_{\mathrm{nn}(t_i)}\right)}^2\right. \\
&\left. +{\left(\dot{E}_{\mathrm{nn}}(t_i)-\left( {\beta}{S_{\mathrm{nn}}(t_i)}{I_{\mathrm{nn}}(t_i)}-{\epsilon}{E_{\mathrm{nn}}(t_i)}\right)\right)}^2\right. \\
&\left. +{\left(\dot{I}_{\mathrm{nn}}(t_i)-\left({\epsilon}{E}_{\mathrm{nn}}(t_i)-{\gamma}{I}_{\mathrm{nn}}(t_i)\right)\right)}^2\right. \\
&\left. +{\left(\dot{R}_{\mathrm{nn}}(t_i)- {\gamma}{I}_{\mathrm{nn}}(t_i)\right)}^2\right. \\
&\left. + 
{\left(S_{\mathrm{nn}}(t_i) + E_{\mathrm{nn}}(t_i) + I_{\mathrm{nn}}(t_i) + R_{\mathrm{nn}}(t_i)-1\right)}^2
\right\}.
\end{split}
\end{align*}
$L_{\mathrm{init}}$ is defined as the sum of the mean squared error of $x_\mathrm{nn}(0)$ and $x_{\mathrm{d}}(0)$.

\section*{Appendix B: Further details on the algebraic observability \cite{meshkat}}\label{sec:meshkat}
According to \cite{meshkat}, the order of the derivatives required to show the existence of such polynomial equations is $N-1$, where $N$ is the dimension of the state vector. Hence, the problem of finding such equations is reduced to eliminating a certain set of variables from a finite set of polynomial equations regarding the derivatives of variables independent of the variables. 
This process can be performed algorithmically based on computer algebra, in particular, using the Gr\"{o}bner basis\cite{iva}. 
Furthermore, computer algebra software, e.g., Singular\cite{singular}, allows the process to be automated. 
An example of a series of Singular commands used to investigate the algebraic observability of $S$ in our target scenario is provided in Appendix \ref{sec:append}. 

\section*{Appendix C: Example of Singular commands} \label{sec:append}
The following is an example of series of commands for Singular \cite{singular} to investigate algebraic observability of $S$ in our target scenario. See Tabel \ref{tab:table} for the notation of variables and parameters of \eqref{eq:seir} in the commands. See Appendices of \cite{robot} for the details of the commands.

\begin{itembox}[l]{Example of Singular commands}
\begin{verbatim}
ring r =(0, b, e, g),(d3I,d3E,d3S,d2I,
d2E,d2S,d1I,d1E,d1S, I, E, S, d3Y, d2Y, 
d1Y, Y),lp;

ideal J = d1S + I*I*b, d2S + I*d1I*b 
+ I*d1S*b, d3S + I*d2I*b + I*d2S*b 
+ 2*d1S*d1I*b, d1E + E*e - I*I*b, 
d2E + d1E*e - I*d1I*b - I*d1S*b, 
d3E + d2E*e - I*d2I*b - I*d2S*b 
- 2*d1S*d1I*b, d1I - E*e + I*g,
d2I - d1E*e + d1I*g, d3I - d2E*e 
+ d2I*g, Y - I, d1Y - d1I,d2Y - d2I,
d3Y - d3I

option(redSB);

groebner(J);
\end{verbatim}
\end{itembox}
\begin{itembox}[l]{Part of the output of the commands}
\begin{verbatim}
(e)*E-d1Y+(-g)*Y 
\end{verbatim}
\end{itembox}
\begin{table}[H]
	\caption{Correspondence of variables. $Y$ denotes the observed variables: $Y = I$.}
	\centering
	\begin{tabular}{ll}
		\toprule
		\cmidrule(r){1-2}
		Variables in \eqref{eq:seir}    & Variables in commands\\
		\midrule
		$S, E, I$ & S, E, I\\
		$\dot{S}, \dot{E}, \dot{I}$ & d1S, d1E, d1I\\
		$\ddot{S}, \ddot{E}, \ddot{I}$ & d2S, d2E, d2I\\
  $\dddot{S}, \dddot{E}, \dddot{I}$ & d3S, d3E, d3I\\
            $Y, \dot{Y}, \ddot{Y}, \ddot{Y}$ & Y, d1Y, d2Y, d3Y\\
		\bottomrule
	\end{tabular}
	\label{tab:table}
\end{table}

\section*{Appendix D: Details of setup for numerical experiments}\label{sec:ne_detail} 
\subsection*{Data preparation}
For the proof of concept, we used artificial data generated by the numerical simulation of \eqref{eq:seir} as the ground truth. 
Specifically, \eqref{eq:seir} is numerically solved over the time domain $[0, 200]$ given the initial states $(S(0), E(0), I(0), R(0)) = (0.99, 0.0, 0.01, 0.0)$ and $(\beta, \epsilon, \gamma) = (0.26, 0.2, 0.1)$. 
The parameter values were selected based on \cite{kuniya}. The Dormand-Prince method with a time-step size $\Delta t = 0.2$ was used as a numerical solver. 
For the input data, $n=50$ observation points are sampled over $[0, 200]$ for bothe training and testing. 
For the training data, the observation points were set at evenly spaced intervals. 
For test data, we randomly sampled data from a uniform distribution over $[0, 200]$. 
As mentioned in Section \ref{sec:prop}, the values of $\dot{I}, \ddot{I}$ at the observed points are required for the proposed framework.
For simplicity, we assumed that these values were provided in advance, leaving an approximation of these values for future work. 
In the following, the values of $\dot{I}$ and $\ddot{I}$ obtained by substituting the numerical solutions of \eqref{eq:seir} into the first and second derivatives of \eqref{eq:seir} are substituted into \eqref{eq:aoeq}. 

\subsection*{Implementation}
For both the baseline and proposed methods, we used the same settings unless otherwise specified. 
For the PINN, we used fully connected neural networks with three hidden layers, of which 50 units preceded the hyperbolic tangent activation function. 
The Glorot uniform was selected to initialize the weight matrix.

In the proposed framework, the objective of the outer loop was to estimate $\epsilon$. In particular, we applied GP-BO using Python scikit-optimize. 
As the acquisition function for GP-BO, we used the Expected Improvement\cite{EI}. We define the minimum test error of learning algebraically observable PINNs as the loss function of GP-BO. 
The number of iterations was set to 30 and the search space was set to $[0.0, 0.5]$.
In the baseline method, the objective of the outer loop is to estimate the initial values of $\epsilon$ to learn the PINN as an inverse problem in the inner loop. 

In the proposed framework, the objective of the inner loop is to predict the trajectories of $S, E, I, R$ through learning algebraically observable PINNs based on \eqref{eq:aoeq}. The PINNs were trained based on adam optimization for 30000 iterations with a learning rate of $10^{-3}$. 
In the baseline method, the objectives of the inner loop are to predict the trajectories of $S, E, I, R$ and estimate the $\epsilon$. 

\section*{Appendix E: Details of the results of numerical experiments}\label{sec:result}






\end{multicols}
\end{document}